\documentclass[letterpaper]{article} 
\usepackage{aaai2026}
\usepackage{times}  
\usepackage{helvet}  
\usepackage{courier}  
\usepackage[hyphens]{url}  
\usepackage{graphicx} 
\usepackage{natbib}  
\usepackage{caption} 
\usepackage{booktabs}
\usepackage{graphicx}
\usepackage{amsmath}
\usepackage{algorithm}
\usepackage{algorithmic}
\usepackage{amsmath}
\usepackage{amssymb}
\usepackage{tabularx}
\usepackage{array}
\usepackage{multirow}
\usepackage{newfloat}
\usepackage{listings}
\usepackage[table]{xcolor}
\DeclareCaptionStyle{ruled}{labelfont=normalfont,labelsep=colon,strut=off} 
\floatstyle{ruled}
\newfloat{listing}{tb}{lst}{}
\floatname{listing}{Listing}
\nocopyright

\title{CoMA: Complementary Masking and Hierarchical Dynamic Multi-Window Self-Attention in a Unified Pre-training Framework}
\author {
    Jiaxuan Li\textsuperscript{\rm 1 \rm 2},
    Qing Xu\textsuperscript{\rm 1 \rm 2},
    Xiangjian He\textsuperscript{\rm 1}\thanks{Corresponding author.},
    Ziyu Liu\textsuperscript{\rm 1 \rm 2},
    Chang Xing\textsuperscript{\rm1 },
    Zhen Chen\textsuperscript{\rm 3},
    Daokun Zhang\textsuperscript{\rm 1},
    Rong Qu\textsuperscript{\rm 2},
    Chang Wen Chen\textsuperscript{\rm 4}
}
\affiliations {
    \textsuperscript{\rm 1} University of Nottingham Ningbo China\\
    \textsuperscript{\rm 2} University of Nottingham UK\\
    \textsuperscript{\rm 3} Yale University\\
    \textsuperscript{\rm 4} The Hong Kong Polytechnic University\\
    \{jiaxuan.li, qing.xu, sean.he, ziyu.liu, chang.xing, daokun.zhang\}@nottingham.edu.cn,\\
    \{scxjl6, scxqx1, Ziyu.Liu1, rong.qu\}@nottingham.ac.uk,\\
    \{zhen.chen\}@yale.edu, \{changwen.chen\}@polyu.edu.hk
}

\usepackage{bibentry}

\begin{document}

\maketitle

\begin{abstract}
Masked Autoencoders (MAE) achieve self-supervised learning of image representations by randomly removing a portion of visual tokens and reconstructing the original image as a pretext task, thereby significantly enhancing pretraining efficiency and yielding excellent adaptability across downstream tasks. However, MAE and other MAE-style paradigms that adopt random masking generally require more pre-training epochs to maintain adaptability. Meanwhile, ViT in MAE suffers from inefficient parameter use due to fixed spatial resolution across layers. To overcome these limitations, we propose the Complementary Masked Autoencoders (CoMA), which employ a complementary masking strategy to ensure uniform sampling across all pixels, thereby improving effective learning of all features and enhancing the model's adaptability. Furthermore, we introduce DyViT, a hierarchical vision transformer that employs a Dynamic Multi-Window Self-Attention (DM-MSA), significantly reducing the parameters and FLOPs while improving fine-grained feature learning. Pre-trained on ImageNet-1K with CoMA, DyViT matches the downstream performance of MAE using only 12\% of the pre-training epochs, demonstrating more effective learning. It also attains a 10\% reduction in pre-training time per epoch, further underscoring its superior pre-training efficiency.

\end{abstract}


\section{Introduction}

In recent years, self-supervised learning has emerged as a prominent paradigm in computer vision research. It learns generalizable feature representations from unlabeled data by designing pretext tasks and demonstrates strong transferability over a wide range of downstream tasks \cite{ermolov2021whitening, oord2018representation, gui2024survey}. Various self-supervised learning methods, including contrastive learning \cite{chen2020simple} and image reconstruction \cite{xie2022simmim}, have demonstrated performance comparable to or exceeding that of traditional supervised pre-training across a variety of downstream tasks such as image classification, object detection, and semantic segmentation, thus highlighting their significant potential for advancing visual representation learning.

\begin{figure}[t]
\centering
\includegraphics[width=1\columnwidth]{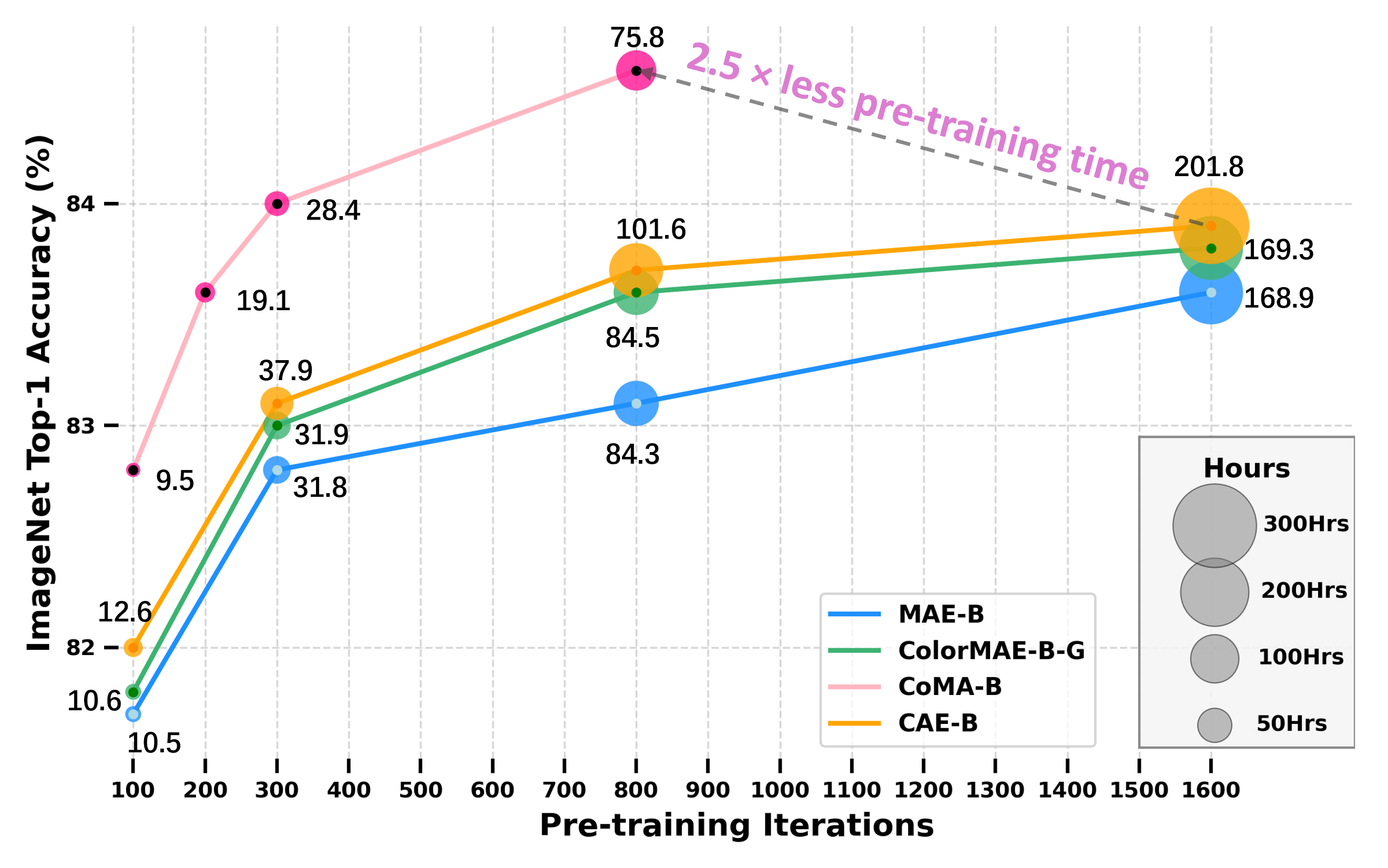}
\caption{Visualization of the relationship between the number of pre-training iterations and ImageNet-1K classification accuracy for the base versions of  MAE \cite{he2022masked}, CAE \cite{chen2024context}, ColorMAE \cite{hinojosa2024colormae}, and CoMA. The radius of each data point represents the total pre-training time, with larger dots indicating longer durations. Specific pre-training hours are annotated beside each point. All models were trained using a 60\% masking ratio.}
\label{performance}
\end{figure}

Masked Autoencoders (MAE) have been extensively explored in recent studies on self-supervised learning, owing to their ability to model only visible tokens through a random masking strategy. This design substantially improves pretraining efficiency, producing a speed-up of 4$\times$ to 10$\times$ in the pre-training phase, while maintaining strong generalization performance across downstream tasks \cite{he2022masked}. Recent studies tend to guide the masking strategy through additional modules and implement random masking within the guided regions. \citep{shi2022adversarial, chen2023improving, krishna2025keypoint}. 

\begin{figure}[t]
\centering
\includegraphics[width=1\columnwidth]{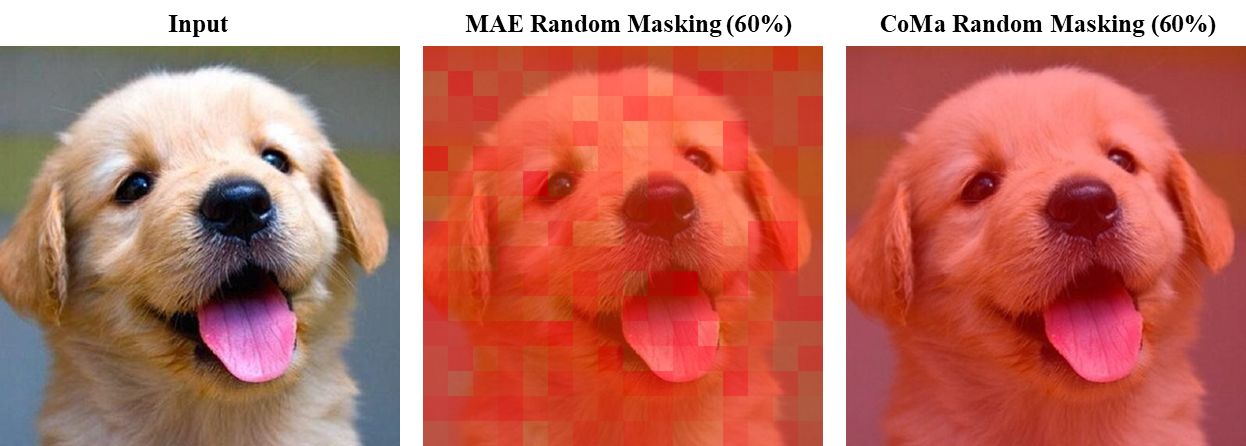}
\caption{Visualization of masking frequency over 1,600 iterations for Random masking (MAE) and Complementary Masking (CoMA). Dark red regions indicate patches with higher masking frequencies, while lighter regions correspond to less frequently masked patches. Both strategies employ a masking ratio of 60\%.}
\label{sampling}
\end{figure}

However, the use of random masking may result in an uneven allocation of sparse supervision signals, wherein certain regions are excessively sampled while others suffer from insufficient supervision. Moreover, introducing additional modules to guide the masking process inevitably incurs increased computational overhead during pre-training. In practice, such sparse supervision signals necessitate additional resampling iterations to ensure that the sampling frequencies of features converge toward statistical uniformity. As shown in Fig.~\ref{sampling}, the sampling statistics of MAE’s random masking over 1,600 iterations reveal that not all patches are uniformly selected as masked tokens, even after extensive pre-training. This uneven distribution may impair the model’s ability to learn from regions with high masking frequencies, and under such conditions, the model may prematurely converge falsely assuming that optimal learning has been achieved under the current masking strategy. Moreover, MAE adopts ViT as its backbone, which partitions images into independent patches and models long-range dependencies. However, this design overlooks local structures and internal correlations within patches, and its fixed spatial resolution further limits adaptability, resulting in low parameter efficiency \citep{yuan2021tokens, ryali2023hiera, liu2021swin, li2025cfformer}.

To enhance both the efficiency and the representational quality of pre-training, we introduce Complementary Masked Autoencoder (CoMA). This approach generates a pair of complementary random masks to ensure that all visual tokens have an equal probability of being selected as visible or masked in each iteration. Based on this, the two complementary sets of visible tokens are fed into the adaptive model and the evaluation model, respectively, which share an identical architecture. Importantly, the parameters of the evaluation model are entirely frozen and updated solely through the guidance of the adaptive model, introducing negligible overhead to pre-training efficiency. This design facilitates uniform feature sampling as shown in Fig.~\ref{sampling}, mitigates the risk of convergence to suboptimal solutions caused by imbalanced masking strategies, ultimately enhancing the model’s adaptability in downstream tasks. Meanwhile, we design a lightweight hierarchical Vision Transformer, DyViT, which integrates the proposed Dynamic Multi-Window Self-Attention to model the relationship between feature points and multi-scale representations, thereby enhancing the perceptual ability of each layer. DyViT significantly reduces the number of parameters and FLOPs, thereby improving the pre-training efficiency of masked autoencoders. We have pre-trained DyViT on ImageNet-1K using the Complementary Masked Autoencoder (CoMA) approach. It converges in fewer epochs and achieves approximately 10\% faster pre-training speed than MAE as shown in Fig.~\ref{performance}. CoMA enables more efficient parameter utilization, facilitates faster convergence, and reduces the model’s dependence on extended pre-training durations. We have evaluated DyViT on multiple downstream tasks, where it demonstrates highly competitive performance. Our main contributions are summarized as follows:
\begin{itemize}
  \item We construct a complementary masking strategy to ensure that all patches receive effective supervision in each pre-training iteration. This approach improves data utilization and mitigates the risk that the model is converging to a local optimum.
  \item We propose DyViT, which adaptively captures multi-scale information to enhance downstream performance. Its hierarchical architecture effectively improves the parameter efficiency of the Vision Transformer.
  \item The proposed Dynamic Multi-Window Self-Attention enables multi-scale modeling within ViT, fundamentally enhancing the model’s ability to perceive features at varying granularities.
  \item CoMA pre-training enables DyViT to achieve superior downstream performance with shorten pre-training time, highlighting its efficiency and effectiveness.
\end{itemize}

\section{Related Work}
\subsection{Masked Autoencoders Representation Learning}
Masked Autoencoders (MAE) draw inspiration from masked language modeling (MLM) in natural language processing, where a portion of input tokens is masked and predicted using the surrounding context \cite{devlin2019bert}. Unlike NLP, where the number of tokens does not affect the computation of self-attention scores due to the inherent design of vision transformers, MAE removes masked patches directly from the input sequence during encoding \cite{he2022masked, feichtenhofer2022masked}. Some recent studies have proposed replacing the random masking strategy in MAE to enhance downstream performance. These approaches mainly involve introducing additional modules \cite{shin2024self, madan2024cl, wang2024rethinking}, or using a teacher-student framework \cite{zhu2024teaching, mo2025dynamic}, which encourages Vision Transformer to focus on regions with high semantic or visual information density during the reconstruction task. Although constraining the masking region using image-informed strategies can improve focus, it still suffers from uneven sampling across patches, often requiring longer training to converge. Additionally, such methods increase models' complexity and computational costs. Enhancing pre-training efficiency while maintaining robustness remains a key challenge.

\subsection{Multi-scale Perception of Vision Transformers}
Multi-scale Vision Transformers effectively capture low-level visual information in the shallow layers and progressively extract more complex semantic features in the deeper layers \cite{fan2021multiscale}. The hierarchical design effectively enhances the model's ability to capture objects of varying sizes within an image, while encouraging the model to maintain robust recognition capability under scale-equivariant conditions \cite{tian2023designing}. Some recent arts demonstrate that the ingenious hierarchical design remains highly effective in Vision Transformers \cite{liu2024vision, ghahremani2024h}. For instance, the Pyramid Vision Transformer (PVT) \cite{wang2021pyramid} employs a hierarchical pyramid architecture to generate multiscale feature representations, thus enhancing the model's representational capacity. Empirical results validate that this hierarchical design facilitates accurate dense predictions for high-resolution inputs in various vision tasks. The Swin Transformer \cite{liu2021swin, liu2022swin} introduces a sliding window mechanism that confines self-attention computation to local windows, thus reducing computational complexity. Hiera, proposed in \cite{ryali2023hiera}, improves downstream task performance through simple yet effective modeling of unit attention, enhancing the model’s ability to perceive local features. Prior studies \cite{long2015fully, liu2016ssd, lin2017feature, zhao2017pyramid, wu2021cvt, zhang2021multi, ghahremani2024h} consistently demonstrate that aggregating information across different granularities significantly boosts model performance on downstream tasks.

\begin{figure}[t]
\centering
\includegraphics[width=1\columnwidth]{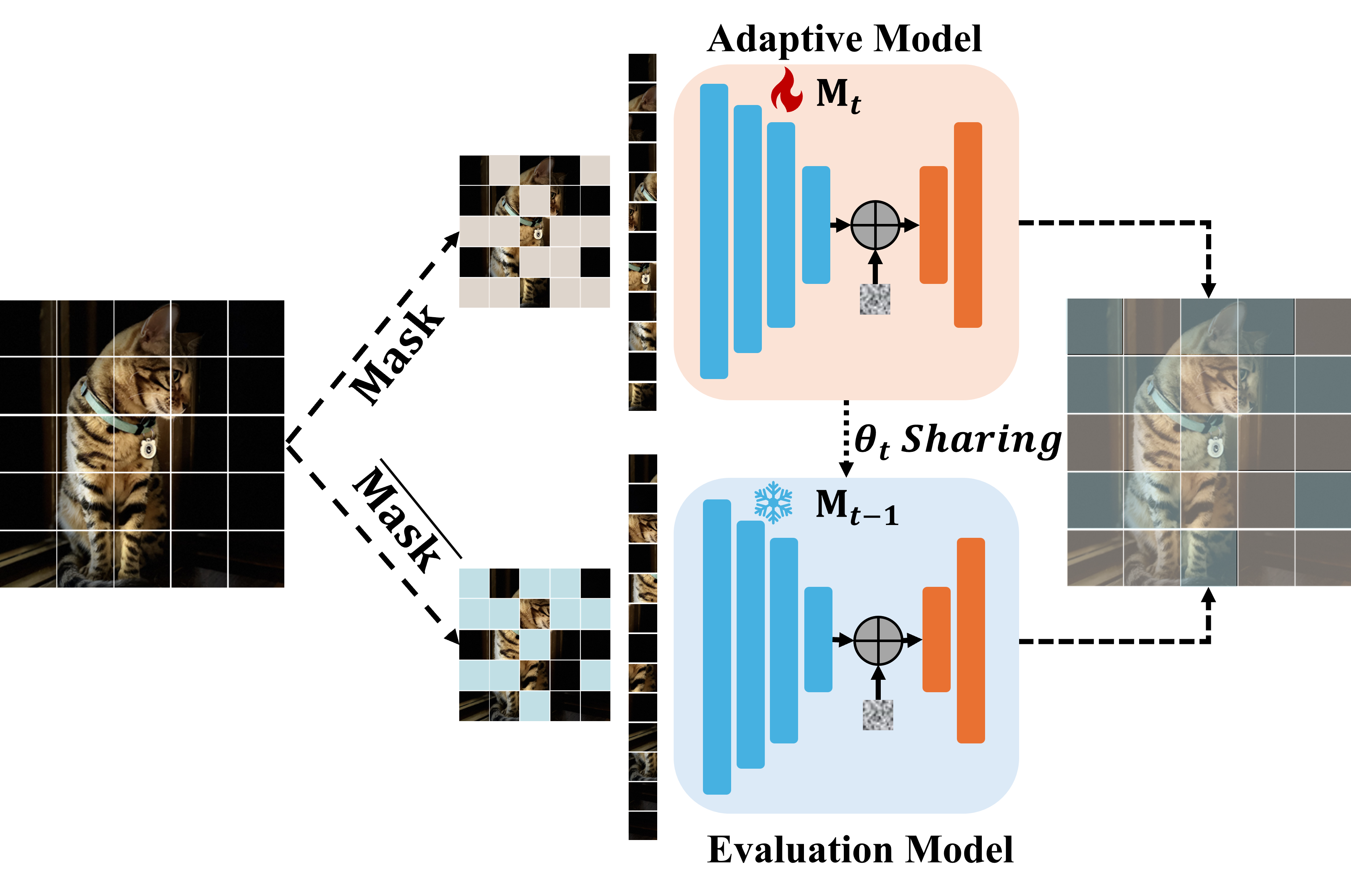}
\caption{\textbf{CoMA: Complementary Masked Autoencoder.} Model $M_t$ serves as the adaptive model and participates in gradient backpropagation, while model $\text{M}_{t-1}$ acts as the evaluation model and remains completely frozen. The parameters of $\text{M}_{t-1}$ are updated solely based on those of $\text{M}_t$ at time step $t$.}
\label{CoMA}
\end{figure}
\section{Methods}
The MAE pre-training process relies on random masking, which results in uneven mask coverage that limits the model’s learning capacity. This misleads the model into believing it has reached an optimal solution, leading to convergence on suboptimal representations derived from insufficient random sampling. Furthermore, the fixed-resolution design of the ViT used in MAE hinders its ability to capture fine-grained features. To address these issues, we introduce CoMA, a novel pre-training framework, alongside DyViT, a hierarchical Vision Transformer, to enhance pre-training efficiency and effectiveness.

\subsection{Complementary Masked Autoencoders}
\begin{figure*}[t]
\centering
\includegraphics[width=\textwidth]{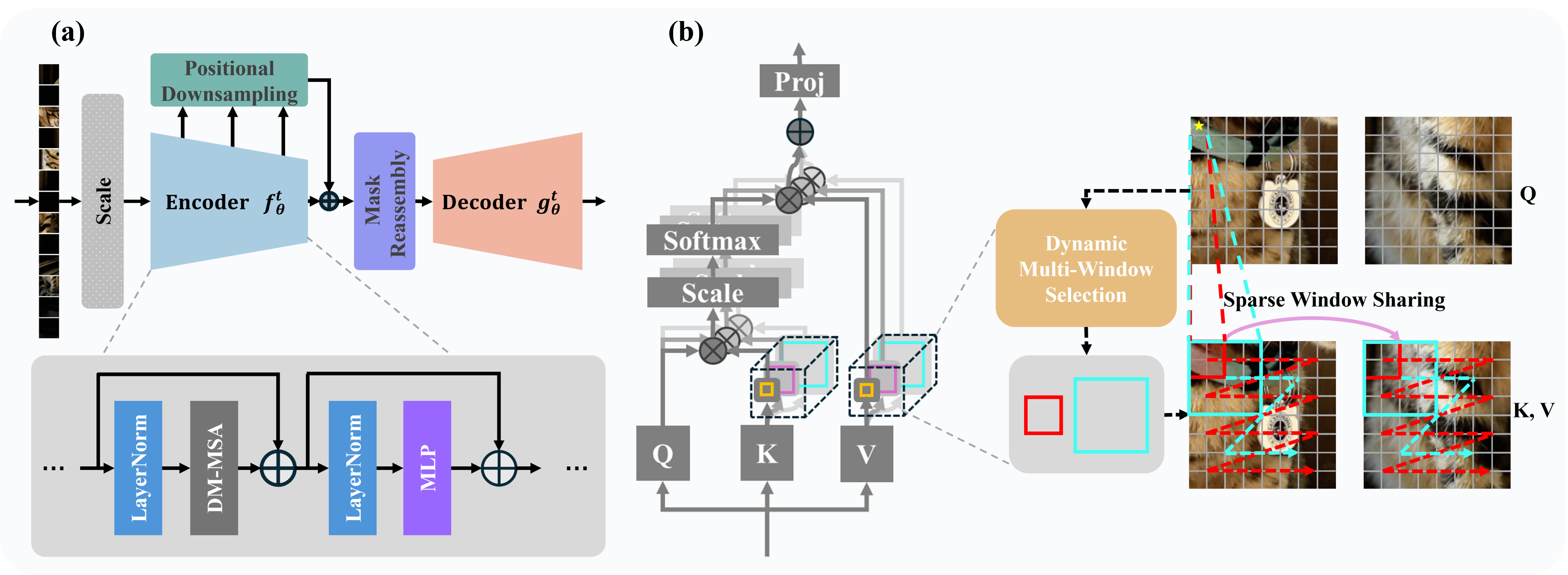}
\caption{The structure of the proposed DyViT model and its core attention mechanism. (a) Overview of the DyViT architecture. (b) Illustration of the Dynamic Multi-window Self-Attention module (DM-MSA).}
\label{DyViT}
\end{figure*}
We propose the Complementary Masked Autoencoder, which employs complementary masking strategies to ensure that the model can predict two mutually exclusive masks with only a single parameter update. This design also maintains the pre-training efficiency.

\noindent\textbf{Complementary Masking.} CoMA adopts a dual-branch complementary masking strategy to ensure that all patches have equal chances of being masked and unmasked within a single epoch, as illustrated in Fig.~\ref{CoMA}, thus promoting more comprehensive feature learning. 

In the two branches, we adopt a complementary random masking strategy to ensure that different subsets of patches are visible in each branch, such that the two masks satisfy the condition \(\text{Mask} + \overline{\text{Mask}} = \mathbf{1}\), where \(\text{Mask}\) denotes the masking strategy used by the adaptive model and \(\overline{\text{Mask}}\) corresponds to that of the evaluation model. Note that both masks are binary matrices, with 1 indicating the positions of preserved vision tokens and 0 representing the tokens to be removed. Given any input image \(\mathbf{X}\), the feature processing proceeds as follows:
\begin{align}
    \text{X}^{\text{M}} = \text{X}\circ\text{Mask}, \quad \text{X}^{\text{C}} = \text{X}\circ\overline{\text{Mask}},
    \label{eq:mask_definition}
\end{align}
where \(\circ\) denotes the Hadamard product. \(\text{X}^{\text{M}}\) denotes the visible tokens retained by \(\text{Mask}\), while \(\text{X}^{\text{C}}\) represents those retained by its complement \(\overline{\text{Mask}}\). If the spatial dimensions of \(\text{Mask}\), \(\overline{\text{Mask}}\), and \(\text{X}\) are inconsistent, interpolation is applied to masks ensure spatial alignment. We remove all tokens with a value of zero and feed the remaining tokens into the encoder. At the end of the encoder, we reinsert empty patches at the positions masked out (i.e., those with zeros in $\text{Mask}$ and $\overline{\text{Mask}}$) to restore the original input structure, following the same procedure as MAE. Finally, the reconstructed features are passed to the decoder, producing \(\text{X}_{\text{rec}}^{\text{M}}\) and \(\text{X}_{\text{rec}}^{\text{C}}\), respectively. Subsequently, we assemble the reconstructed patches into a complete image, which corresponds to a full-image reconstruction. The operation is performed by:
\begin{align}
    \text{X}_{\text{rec}} = \text{X}_{\text{rec}}^{\text{M}}\circ \overline{\text{Mask}} + \text{X}_{\text{rec}}^{\text{C}}\circ \text{Mask},
\end{align}
where $\text{X}_{\text{rec}}$ is the final reconstructed features with the same shape as $\text{X}$. 
The final reconstructed full images are supervised using the Mean Squared Error (MSE) loss, formulated as:
\begin{align}
    \mathcal{L}_{\text{MSE}} = \Vert\text{X}_{\text{rec}} - \text{X}\Vert_F^2.
\end{align}
\noindent\textbf{Efficient Pre-training.} In CoMA, the parameters of the evaluation model \( \theta_{\text{Evaluation}} \) are frozen to improve the pre-training efficiency. These parameters are directly updated from the adpative model \( \theta_{\text{Adaptive}} \) in the previous pre-training step. This strategy, inspired by the Double Deep Q-Network \citep{van2016deep}, enhances the stability of pre-training. As training progresses, the predictions of \( \theta_{\text{Evaluation}} \) are expected to approximate those of \( \theta_{\text{Adaptive}} \). The parameter update is performed as:
\begin{align}
\theta_{\text{Evaluation}}^{(t)} \leftarrow \theta_{\text{Adaptive}}^{(t-1)}.
\end{align}
We employ parameter sharing ``$\leftarrow$'' instead of an EMA strategy \cite{tarvainen2017mean} because the Evaluation Model functions merely as an auxiliary branch that acts as a shadow of the Adaptive model to assess its performance under a complementary masking scheme. This pre-training strategy enables dense supervision over all patches within the image, effectively alleviating the sampling bias introduced by random masking and ensuring more uniform coverage across the spatial domain. 

We freeze the Evaluation Model for two primary reasons. Firstly, we aim to improve the efficiency of pre-training. Compared with contrastive learning \cite{chen2020simple}, only the Adaptive model needs to be trained after freezing the parameters, which significantly reduces the computational cost. Secondly, because the Evaluation Model acts solely as a projection of the Adaptive Model and is responsible only for predicting the Adaptive Model’s output under the complementary mask, the temporal offset between them contributes to the stability of the improved model. This mechanism aligns with the principles of Double Deep Q-Network \cite{van2016deep}. Moreover, it enables parallel inference on the input data, avoiding the inefficiency of sequential reprediction.

\subsection{Hierarchical Dynamic Vision Transformer}
Hierarchical Vision Transformers progressively increase channel depth while reducing spatial resolution to capture high-level semantics. This structure improves parameter efficiency, enables multi-scale feature extraction, and reduces overall complexity. Building upon this idea, we design DyViT, a framework that can be seamlessly integrated into the CoMA pre-training paradigm and other MAE variants.\\
\\
DyViT consists of four layers, with spatial resolutions progressively downsampled by factors of 4×, 8×, 16×, and 32× relative to the original input size. In the \textit{scale} layer, we apply a convolution operation with a kernel size of $7\times7$, stride of $4\times4$ and padding of $3\times3$ to downsample the feature maps by a factor of $4\times$. We then apply patch-level masking on the feature maps, consistent with Equation~(\ref{eq:mask_definition}), followed by the removal of the masked patches. All other layers employ a $2\times2$ max pooling operation within each patch to achieve the reduction in spatial resolution. In the first two layers, we employ multiple sets of DM-MSA blocks to model spatial correlations between pixels and multi-scale feature descriptors extracted with different window sizes, as shown in Fig.~\ref{DyViT}(a). This not only reduces the number of model parameters, but also facilitates effective modeling of spatial relationships across scales. In the deeper layers (i.e., layers 3 and 4), we adopt global self-attention \cite{dosovitskiy2020image} to capture long-range dependencies and enhance high-level semantic understanding.

As illustrated in Fig.~\ref{DyViT}(a), the \textit{Positional Downsampling} module is designed to extract features from the outputs of different layers and fuse them together to capture multi-scale information. The output features \( \text{X}_i \) from the four layers input to this module have the following shapes:
\begin{equation}
\begin{aligned}
\text{X}_i &\in \mathbb{R}^{B \times f n p_{i}^2 \times C_i}, \; i \in \{1, 2, 3, 4\}, \;p_i \in \{8, 4, 2, 1\},\\
\end{aligned}
\end{equation}
where \( B \) is the batch size, \( n \) the number of patches, \( f \) the masking ratio, and \( C_i \) the channel dimension.

For the output features \( \text{X}_i \) from four layers, we perform patch downsampling using strided convolution. The detailed operation is as:
\begin{equation}
\begin{aligned}
\text{Y}_i = \text{Conv}_{k, s}(\text{Y}_{i-1}) + \text{X}_{i+1}, \;\; i \in \{1, 2,3\},
\end{aligned}
\end{equation}
where \( \text{Y}_{i-1} \) is initialized as \( \text{X}_1 \), and both the kernel size \( k \) and the stride \( s \) are set to \( 2^{4 - i} \). After hierarchical fusion, the output feature \( \text{X}_{\text{out}} = \text{Y}_3 \) is obtained, which has the shape \( \mathbb{R}^{B \times n \times C_4} \). The fused representation is subsequently processed by the \textit{Mask Reassembly} module. Specifically, noise patches are inserted at the positions corresponding to the masked regions. The resulting latent features are then passed to a decoder composed of 8$\times$ global attention blocks \cite{dosovitskiy2020image} for reconstruction.
\subsection{Dynamic Multi-Window Self-Attention}
Our design facilitates attention interactions between individual feature points and region-level features extracted from multiple window scales. Importantly, our attention design allows information exchange across patches, thereby enhancing global context modeling at shallow layers while maintaining computational efficiency during both training and inference. 

Building upon the strength of self-attention in capturing global dependencies, we apply the standard query, key and value projections to the input feature \( \text{X} \) in the encoder.  For any input $\text{X} \in \mathbb{R}^{B \times np^2 \times C}$, the initial projection of features into the query (Q), key (K), and value (V) representations is defined by:
\begin{equation}
    \text{Q} = \text{X} \text{W}^Q, \quad \text{K} = \text{X} \text{W}^K, \quad \text{V} = \text{X} \text{W}^V,
\end{equation}
where $\text{W}^Q, \text{W}^K, \text{W}^V\in \mathbb{R}^{C \times C}$. To capture multi-scale information at a specific resolution, our Dynamic Multi-Window Selection module first determines candidate convolutional kernel sizes, strides, and the number of branches based on the patch size as shown in Fig.~\ref{DyViT}(b). The operation of the Dynamic Multi-Window Selection module can be formulated as:
\begin{equation}
    \mathcal{K} = \left\{ k_i \;\middle|\; k_i = \frac{p}{2^i},\; i \in \mathbb{N},\; 2^i \mid p \right\},
\end{equation}
where \(k_i \in \mathcal{K}\) denotes the \(i\)-th possible kernel size and p is the patch size. Based on each kernel size in set $\mathcal{K}$, we apply stride convolution to extract features in a sparse manner. As illustrated in Fig.~\ref{DyViT}(b), the convolutional kernel operates strictly within the local region of each patch, capturing localized information. To enable the model to capture multi-scale features, we further incorporate a self-attention mechanism to model spatial relationships across positions and scales. The operation is formulated as:
\begin{equation}
\text{Attn}_k(\text{Q}, \text{K}, \text{V}) = \text{Softmax}\left( \frac{\text{Q}\left( \text{Conv}_{k}(\text{K}) \right)^\top}{\sqrt{d}} \right) \text{Conv}_{k}(\text{V}),
\end{equation}
where \( \mathrm{Conv}_k(\cdot) \) denotes a strided convolution with both kernel size and stride equal to \(k \in \mathcal{K}\). Due to the inherently sparse sampling nature of strided convolution, a single-scale window lacks the capacity to capture correlations between adjacent features. In contrast, a multi-scale strided convolutional design effectively addresses this limitation. Larger kernels offer broader receptive fields that subsume those of smaller ones, enabling the recovery of local feature correlations that may be overlooked at finer scales. Finally, the multi-scale features are integrated and linearly projected to enhance the model’s ability to capture information over different levels of granularity. The procedure is as:
\begin{equation}
    \text{X}_{out} = \left[\sum_{k \in \mathcal{K}} \text{Attn}_k(\text{Q}, \text{K}, \text{V})\right]\text{W}_{out},
\end{equation}
where $\text{W}_{\text{out}}$ denotes the linear projection with shape $\mathbb{R}^{C \times C}$.

\section{Experimental Results}
\subsection{Implementation Details}
\begin{table}[h]
\centering
\small
\begin{tabular}{c|c|c|c}
\hline
\textbf{Model} & \textbf{\#Channels} & \textbf{\#Blocks} & \textbf{\#Heads} \\
\hline
DyViT-S & [96-192-384-768] & [1-2-11-2] & [2-4-8-16]\\
DyViT-B & [112-224-448-896] & [2-3-16-3] & [2-4-8-16]\\
\hline
\end{tabular}
\caption{DyViT Configurations. \#Channels represents the number of channels in each layer, \#Blocks denotes the number of Transformer blocks per layer, and \#Heads refers to the number of attention heads in the multi-head attention mechanism for each layer.}
\label{tab:LiteViT_config}
\end{table}
We perform pre-training on the ImageNet-1K training set using 4$\times$ H200 GPUs, each equipped with 141 GB of memory. The model is pre-trained with a batch size of 4096 using a patch size of 32×32 and a masking ratio of 60\% for the Adaptive model and 40\% for the Evaluation model. On average, every 100 epochs take approximately 9.5 hours for DyViT-B. AdamW is employed as the optimizer with a batch size of 4096. For downstream tasks with resolutions differing from pre-training, positional embeddings are reinserted prior to fine-tuning. Only the encoder weights are transferred. In image classification, global average pooling followed by a single-layer MLP maps the features to the target label space. The configurations of DyViT-Small and DyViT-Base are detailed in Table~\ref{tab:LiteViT_config}.
\subsection{Model Evaluation}
\begin{table*}[t]
  \centering
  \begin{tabularx}{\textwidth}{>{\raggedright\arraybackslash}X c c c c c c}
      \hline
      \textbf{PT-Method} & \textbf{PT-Task} & \textbf{Arch.}& \textbf{Epochs} & \textbf{Acc.} & \textbf{FLOPs} & \textbf{Params} \\
      \hline
      DINO \cite{caron2021emerging} & CL & ViT-S & 3200 & 82.0 & 5 & 22\\
      BEiT \cite{bao2021beit} & MIM & ViT-S & 800 & $81.4^\ddagger$ & 5 & 22\\
      MAE \cite{he2022masked} & MAE & ViT-S & 800 & $81.6^\ddagger$ & 5 & 22\\
      SimMiM \cite{xie2022simmim} & MIM & Swin-S  & 1600 & 83.2 & 9 & 50\\
      iBOT \cite{zhou2021ibot} & MIM+CL & ViT-S & 3200 & 82.3  & 5 & 22 \\
      \hline
      \rowcolor{gray!10}
      DyViT & CoMA & DyViT-S & 300 & 83.6 & 6 & 35 \\
      \rowcolor{gray!10}
      DyViT & CoMA  & DyViT-S & 800 & \textbf{83.9} & 6 & 35 \\
      \hline
      DINO \cite{caron2021emerging} & CL & ViT-B & 1600 & 83.6 & 18 & 87\\
      BEiT \cite{bao2021beit}& MIM & ViT-B  &  800 &  83.2 & 18 & 87 \\
      MAE \cite{he2022masked}& MAE & ViT-B  &  1600 & 83.6 & 18 & 87 \\
      iBOT \cite{zhou2021ibot} & MIM+CL & ViT-B & 1600 & 84.0 & 18 & 87\\
      SimMiM \cite{xie2022simmim} & MIM & Swin-B & 1600 & 83.8 & 15 & 88\\
      $\text{A}^2$MIM \cite{li2022architecture} & MIM & ViT-B &  800 & 84.3 & 18 & 87\\
      Hybrid Distill \cite{shi2023hybrid} & Distill-MAE & ViT-B &  300 & 83.7 & 18 & 87\\
      CAE \cite{chen2024context} & MIM & ViT-B &  1600 & 83.9 & 18 & 87 \\
      ColorMAE    \cite{hinojosa2024colormae} & MAE &  ViT-B &  1600 & 83.8 & 18 & 87\\
      DeepMIM-MAE  \cite{ren2025deepmim} & Deep-supervision + MAE & ViT-B & 1600 & 84.0 & 18 & 87\\
      \hline
      \rowcolor{gray!10}
      DyViT & MAE & DyViT-B &  300 & 83.6 & 12 & 70 \\
      \rowcolor{gray!10}
      DyViT & CoMA & DyViT-B &  300 & 83.9 & 12 & 70 \\
      \rowcolor{gray!10}
      DyViT & MAE &  DyViT-B & 800 & 84.2 & 12 & 70\\
      \rowcolor{gray!10}
      DyViT & CoMA & DyViT-B &  800 & \textbf{84.6} & 12 & 70 \\
      \hline
    \end{tabularx}%
  \caption{Comparison of Top-1 Accuracy (\%) between DyViT and state-of-the-art models on Imagenet-1K dataset. All models are evaluated with a resolution of $224\times224$ under consistent downstream configurations. PT-Task refers to the pre-training methods. Bold indicates the best performance. Results marked with $\ddagger$ are sourced from \cite{zhou2021ibot}.}
  \label{tab:classification_performance}
\end{table*}
\textbf{Image Classification.} We present a comprehensive performance comparison among self-supervised Vision Transformers, as summarized in Table~\ref{tab:classification_performance}. Leveraging the CoMA pre-training framework, DyViT consistently outperforms a range of competitive architectures. Remarkably, when compared to the MAE's pre-trained Vision Transformer, DyViT pre-trained under CoMA achieves superior results using 300 epochs, highlighting its exceptional pre-training efficiency. 

To further substantiate this observation, we conduct a direct comparison between DyViT models pre-trained with MAE and CoMA. At 800 training epochs, the DyViT model pre-trained with CoMA outperforms MAE's pre-trained counterpart by +0.4, reaffirming the superior pre-training effectiveness afforded by the CoMA framework. Notably, under the same pre-training budget, CoMA's pre-trained DyViT attains a top-1 classification accuracy of 84.6\%, surpassing numerous existing state-of-the-art approaches. These findings collectively underscore the significant advantages of CoMA in enhancing effectiveness of the pre-training process. 

Moreover, DyViT exhibits notable architectural efficiency. Compared to ViT-B operating at the same input resolution, DyViT-B achieves a reduction of approximately 20\% in the number of parameters and a decrease of 33\% in FLOPs. Despite the reductions in both number of parameterss and FLOPs, DyViT achieves superior performance, highlighting its effective parameter utilization.

\noindent\textbf{Semantic Segmentation.} For semantic segmentation on the ADE20K dataset \cite{zhou2017scene}, DyViT follows the same fine-tuning protocol as MAE, employing UperNet as the decoder \cite{xiao2018unified}. Table~\ref{tab:ade20k} summarizes the performance of DyViT compared to state-of-the-arts approaches on this benchmark. Remarkably, our model achieves 51.5 mIoU with only 800 pre-training epochs, significantly outperforming existing state-of-the-art (SOTA) methods, and outperforms MAE and ColorMAE by 3.4\% and 2.2\%, respectively, on the mIoU metric. These results strongly validate the effectiveness of the proposed hierarchical design in improving the model’s ability to capture fine-grained semantic representations. Furthermore, DyViT captures visual tokens at different scales with a dynamic multi-window strategy, modeling and fusing features extracted from keypoints and windows at multiple scales to enhance perception across multiple granularities.
\begin{table}[h]
  \centering
  \resizebox{\linewidth}{!}{ 
    \begin{tabular}{l|c|c|cc}
      \hline
      \multirow{2}{*}{\textbf{PT-Task}} & \multirow{2}{*}{\textbf{Epochs}} & \multicolumn{1}{c|}{\textbf{ADE20K}} & \multicolumn{2}{c}{\textbf{COCO}} \\
      \cline{3-5}
      & & \textbf{mIoU} & \textbf{AP$^{\text{box}}$} & \textbf{$\text{AP}^{\text{mask}}$} \\
      \hline
      DINO \cite{caron2021emerging} & 400 & 47.2 & 46.8 & 41.5 \\
      SdAE \cite{chen2022sdae} & 300 & 48.6 & 48.9 & 43.0 \\
      MAE \cite{he2022masked}& 1600 & 48.1 & 50.3 & 44.9\\
      $\text{A}^2$MIM \cite{li2022architecture}& 800 & 49.0 & 49.4 & 43.5 \\
      Hiera \cite{ryali2023hiera} & 1600 & 50.8 & 52.2 & 46.3 \\
      ColorMAE \cite{hinojosa2024colormae} & 1600 & 49.3 & 50.1 & 44.4 \\
      CAE \cite{chen2024context} & 1600 & 50.2 & 50.2 & 44.2 \\
      \hline
      \rowcolor{gray!10}
      CoMA & 800 & \textbf{51.5} & \textbf{53.1} & \textbf{46.5} \\
      \hline
    \end{tabular}
  }
  \caption{Performance comparison on ADE20K~\cite{zhou2017scene} for semantic segmentation and COCO~\cite{lin2014microsoft} for object detection and instance segmentation. All models are pre-trained on ImageNet-1K and use the base version. Input resolutions are $512 \times 512$ for ADE20K \cite{zhou2017scene} and $768 \times 768$ for COCO with Mask R-CNN \cite{he2017mask}.}
  \label{tab:ade20k}
\end{table}

\noindent\textbf{Object Detection and Instance Segmentation.} We fine-tune Mask R-CNN~\cite{he2017mask} in an end-to-end manner on the COCO dataset \cite{lin2014microsoft} and report both $\text{AP}^{\text{box}}$ for object detection and $\text{AP}^{\text{mask}}$ for instance segmentation. Our implementation follows the ViT-Det training methods~\cite{li2022exploring}, while integrating multi-scale features extracted by DyViT into a Feature Pyramid Network (FPN)~\cite{lin2017feature}. As shown in Table~\ref{tab:ade20k}, our model surpasses the SOTA method, Hiera-B~\cite{ryali2023hiera}, by 0.9 in AP\textsuperscript{box} and 1.2 in AP\textsuperscript{mask}. Compared to MAE \cite{he2022masked}, our approach achieves a notable improvement of 2.8 in AP\textsuperscript{box} and 2.6 in AP\textsuperscript{mask}. DyViT’s hierarchical structure and DM-MSA’s coarse-to-fine modeling enable more robust object detection and instance segmentation.

\begin{table}[h]
\begin{center}
\footnotesize
\resizebox{\linewidth}{!}{
\begin{tabular}{l|c|c}
\hline
\textbf{Method} & \textbf{Epochs} & \textbf{Top-1 Acc.} \\
\hline
 BEiT \cite{bao2021beit} & 800 & 83.2 \\
 MAE \cite{he2022masked} & 1600 & 83.6 \\
 MultiMAE \cite{bachmann2022multimae} & 1600 & 83.3 \\
 SemMAE \cite{li2022semmae} & 800 & 83.3 \\
 ColorMAE \cite{hinojosa2024colormae} & 1600 & 83.8 \\
\hline
\rowcolor{gray!10}
 CoMA (Ours) & 800 & \textbf{84.1} \\
\hline
\end{tabular}
}

\end{center}
\caption{Comparison of CoMA with other masked modeling methods using ViT-Base. All methods are pre-trained and fine-tuned on ImageNet-1K with 224$\times$224 resolution.}
\label{tab:CoMA_ablation}
\end{table}

\begin{figure*}[t]
\centering
\includegraphics[width=\textwidth]{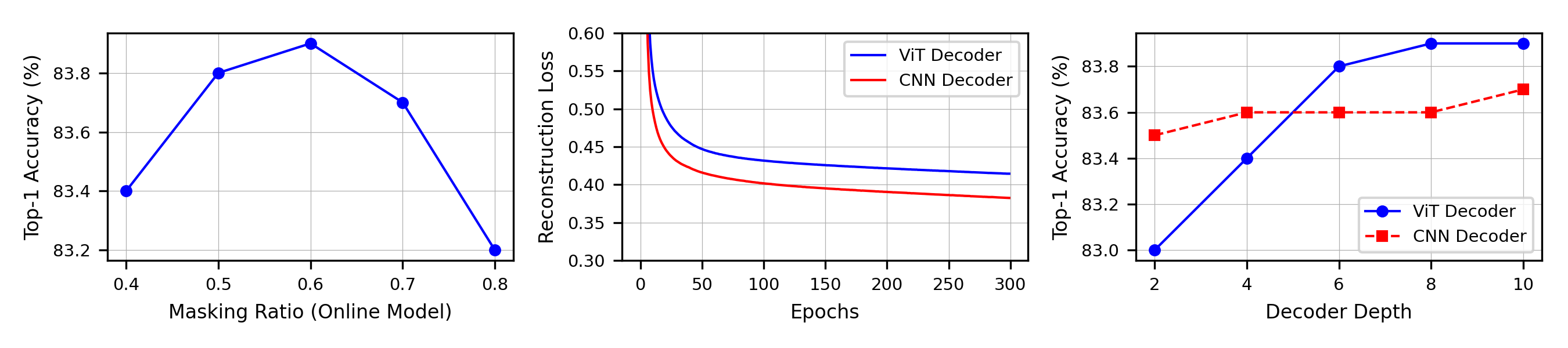}
\caption{Ablation study across masking ratios with $32 \times 32$ patches and 300 pre-training epochs (left); reconstruction loss on the validation set using an 8-layer decoder (middle); impact of decoder depth on classification transferability (right).}
\label{ratioVSacc}
\end{figure*}
\subsection{Ablation Study}
\textbf{Effectiveness of CoMA.} We evaluate the image classification performance of masked autoencoder frameworks, as presented in Table~\ref{tab:CoMA_ablation}. To ensure a fair comparison, ViT-B is adopted as the unified backbone \cite{dosovitskiy2020image}. According to the evaluation results, ViT-B pre-trained with the CoMA framework achieves a top-1 accuracy of 84.1\% on ImageNet-1K, surpassing other MAE-based methods with the same or fewer pre-training epochs. This shows that CoMA facilitates faster model convergence. CoMA's complementary masking strategy significantly enhances pre-training efficiency by enabling more informative and comprehensive sampling across all patches, thereby improving data utilization and effectively reducing the overall pre-training duration.

\noindent\textbf{Masking Ratio.} In MAE \citep{he2022masked}, the $16\times16$ patch division benefits high masking ratios, such as 75\%, by allowing for finer learning of local textures and structures. Smaller patches promote more dispersed visible patches, enhancing the encoder’s capacity with diverse contextual information. However, compared to $16\times16$, a patch size of $32\times32$ can improve pre-training efficiency more effectively, as it results in fewer tokens to process. Meanwhile, since DyViT applies a 4$\times$ downsampling in the first layer to filter out redundant information, using larger patches is more suitable for our architecture, ensuring that each patch contains sufficient information to support the modeling of internal features through Dynamic Multi-Window Self-Attention. 

Setting 300 epochs as the baseline, we systematically evaluate the impact of different masking ratios, with the results presented in Fig.~\ref{ratioVSacc}. Empirical results indicate that a masking ratio of 60\% yields the best performance for the adpative model. In contrast, excessively high masking ratios lead to a notable decline in accuracy. As larger patches encapsulate more information, the corresponding masked regions obscure significantly more features, covering up to four times the content of a patch of $16 \times 16$. Consequently, excessive masking impairs semantic understanding and learning efficiency. Our model’s multi-scale feature modeling within each patch effectively mitigates limitations of larger patch sizes.

\noindent\textbf{Decoder Depth.} We evaluate two decoder architectures: ConvNeXt \cite{liu2022convnet} and ViT \cite{dosovitskiy2020image}. The results show that the ConvNeXt decoder achieves superior reconstruction performance on the ImageNet-1K validation set, demonstrating lower loss values and faster convergence (see Fig.~\ref{ratioVSacc}). This advantage is primarily attributed to the stronger inductive bias inherent in CNN-based decoders.

However, reconstruction loss alone does not fully reflect the generalization capacity of the pre-trained model. In downstream ImageNet-1K classification tasks, increasing the decoder depth to 6 layers enables DyViT pre-trained with a ViT decoder to outperform its counterpart pre-trained with a ConvNeXt decoder. Moreover, the performance of the ViT decoder tends to converge between 8 and 10 layers. Based on these observations, we adopt a decoder composed of 8 ViT blocks in our model.

\noindent\textbf{Pre-training Efficiency.} We conduct pre-training on DyViT, MAE~\cite{he2022masked}, CAE~\cite{chen2024context}, and ColorMAE~\cite{hinojosa2024colormae}, and compare their pre-training efficiency under the same masking ratio, as shown in Fig.~\ref{performance}. Our model demonstrates superior pre-training efficiency compared to state-of-the-art methods, achieving approximately 10\% faster training speed than MAE \cite{he2022masked} and ColorMAE \cite{hinojosa2024colormae}, while requiring only half the number of pre-training epochs to significantly surpass their performance. In particular, it outperforms CAE's 1600-epoch results with only 800 epochs. In downstream tasks, it achieves +2.9 AP$^\text{box}$ over CAE \cite{chen2024context} in object detection and segmentation (Table~\ref{tab:ade20k}), and improves top-1 accuracy by +1.0 and +0.8 over MAE and ColorMAE, respectively, in ImageNet-1K classification, while using fewer parameters and FLOPs.

\section{Conclusion}
Efficient and effective masked autoencoders are increasingly essential for better utilization of limited computational resources, since principled and adaptive masking strategies can significantly reduce pre-training time and improve downstream task performance. In this work, we propose CoMA, which provides richer complementary information to enhance representation learning, alongside DyViT, an efficient multi-scale vision transformer that improves parameter efficiency and accelerates pre-training, while maintaining strong performance on diverse complex natural image tasks. We hope that our exploration will inspire new perspectives and contribute to future advances in the vision community.

\bibliography{aaai2026}

\end{document}